# Efficient Pedestrian Detection in Top-View Fisheye Images Using Compositions of Perspective View Patches

[1]Sheng-Ho Chiang, [1]Tsaipei Wang[*], and [2]Yi-Fu Chen

[1]Department of Computer Science, National Chiao Tung University, Hsinchu, Taiwan, ROC

[2]Chunghwa Telecom Laboratories, Taoyuan, Taiwan, ROC

[*] Corresponding Arthur (E-mail: wangts@cs.nctu.edu.tw)

## Abstract

Pedestrian detection in images is a topic that has been studied extensively, but existing detectors designed for perspective images do not perform as successfully on images taken with top-view fisheye cameras, mainly due to the orientation variation of people in such images. In our proposed approach, several perspective views are generated from a fisheye image and then concatenated to form a composite image. As pedestrians in this composite image are more likely to be upright, existing detectors designed and trained for perspective images can be applied directly without additional training. We also describe a new method of mapping detection bounding boxes from the perspective views to the fisheye frame. The detection performance on several public datasets compare favorably with state-of-the-art results.

# I. Introduction

Top-view (ceiling-mounted) fisheye cameras are widely used in visual surveillance applications. Compared with perspective cameras, what makes fisheye cameras attractive is their very large viewing angles, allowing the coverage of a large space using a single camera. Another benefit of top-view cameras is reduced occlusion among objects in a scene.

The main challenge of pedestrian detection in top-view fisheye images is two-fold: First, people's orientations mostly point outward from the image center instead of being upright. Secondly, their appearances vary with their distances to the image center. Both pose difficulties for pedestrian detectors trained with perspective images.

Compared with the enormous body of research on pedestrian detection for perspective cameras, studies on pedestrian detection for top-view fisheye cameras have been relatively scarce. One possible reason is the lack of publicly available datasets, and it is also inconvenient for researchers to do data collections themselves. While the idea of transforming regular perspective images into fisheye views for training such detectors has been proposed previously [24], this approach is not applicable to top-view fisheye images due to the drastically different viewpoints. Only until very recently that some public datasets of top-view fisheye images useful for pedestrian detection become available [1,21,23] and allow for comparisons between techniques.

Early works on pedestrian detection in top-view fisheye images are mostly geared toward tracking, as surveillance or smart home is their main application scenario. Here the data are in the form of videos from fixed cameras, naturally leading to approaches that just extract and track foreground blobs [14,20]. Information from foreground segmentation is also used in some later pedestrian detectors [22,34]. More sophisticated and successful features for pedestrian detection, such as HOG (Histogram of Oriented Gradients) [5] and ACF (Aggregate Channel Features) [8], have also been used with fisheye images [4,6,13,34]. More recently, several of the very popular CNN (Convolutional Neural Networks) based object detectors (most common ones being YOLO [25] and Mask-RCNN [11]) have been adopted as well [15,22,31,32,35].

In order to apply pedestrian detection techniques for perspective images to fisheye images, we need to solve the problems caused by the two aforementioned challenges. Regarding the orientation variation of people in such an image, the problem is that typical detectors are trained mainly to detect people that appear approximately upright. The most popular solution is to transform part of the fisheye image in a way that makes the people in the image appear approximately upright. Techniques of this class include those in [3,31], which differ with one another in terms of how the sub-image is selected and transformed. However, such techniques can be computationally expensive if the transform and detection steps have to be applied to a large number of sub-images. A related approach is to transform a fisheye image into a single 360-degree panoramic image for pedestrian detection [12,16]. People that are not directly under the camera (i.e., located in the peripheral region in the fisheye image) will appear upright, but those more directly under the camera (i.e., located in the central region in the fisheye image) will be too badly distorted for detection. The transform of image features have also been studied [4,13]. More recently, researchers have also attempted direct training or transfer learning of convolutional neural network (CNN) based detectors specifically for pedestrian detections in fisheye images [22,32,35].

The second major challenge, the appearance variation along the radial direction of the image, has received less attention. Example works include [29], which uses different silhouette templates to match to foreground regions at different positions, and [6,34], which attempt to train different classifiers (to distinguish people from other objects) for different regions.

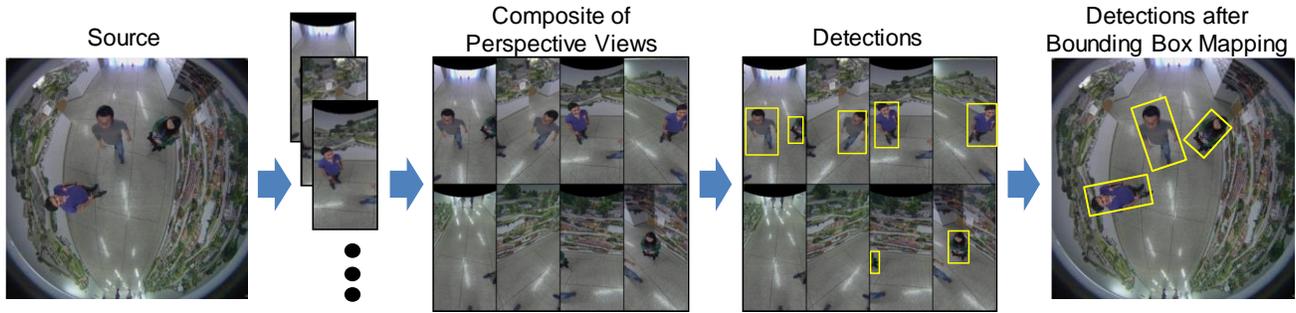

Fig. 1. The processing steps of our pedestrian detection system.

In this paper we propose a new technique of pedestrian detection in fisheye images that is both efficient and flexible. Fig. 1 describes the design of our system. The idea is that we form several perspective views from one fisheye image and, instead of performing detection in these views separately, we combine them into a square composite image, on which the pedestrian detection is performed. This allows us to have the good of two approaches: First, pedestrians in the composite image are approximately upright in the perspective views, hence allowing popular pretrained detectors like YOLO to detect them effectively without further training. Secondly, the detection step only needs to be applied once. This is much more efficient than previous methods such as [31], which incurs high computational cost because the detection step is applied to each perspective view separately.

Our system outputs rotated rectangular bounding boxes in the fisheye frame. As the pedestrians in the image exhibit different orientations, rotated bounding boxes tend to fit them better than axis aligned bounding boxes that are normally used [32]. This requires a post-processing step to transform detected bounding boxes in the perspective views to the fisheye frame. We will call this step Bounding Box Mapping. This mapping is achieved through bounding box regression, which is explained in detail in Section III-D.

Overall, the innovations and contributions of this paper include the following:

(1) We propose to perform pedestrian detection on composite images of perspective views, so that pretrained detectors can be used directly with little loss of efficiency.

(2) A regression based bounding box mapping method to transform bounding boxes from perspective views to the fisheye frame, including the analysis of non-maxima suppression (NMS) methods for integrating detections from multiple perspective views.

Previous researchers tend to use the extra computational cost of warping (from fisheye images to perspective views) as a reason against using perspective views. However, our experimental results indicate that the cost of this step is insignificant compared to the cost of the detector itself. The detection accuracy also compares favorably with state-of-the-art methods.

## II. Related Works

Among existing works on pedestrian detection in top-view fisheye images, [14] is an early one that only handles very simple environments with a person treated as a blob after foreground segmentation. The work in [20] also treated blobs in foreground as detections of people, and proposed to track the detected person using Kalman filtering, creating the effect of automatic virtual perspective PTZ cameras. Foreground segmentation is applied to the panorama of the fisheye image in [12]. To take into account appearance variations at different radial distances, probabilistic shape masks pre-computed from training images are used in [29], with the assumption that the people are always upright and standing, facing toward or away from the camera. This is also the first proposed method that can handle occlusions between people. A network of fisheye cameras is presented in [33] for multi-object tracking with detections also coming from foreground blobs.

Following the success of HOG features in pedestrian detection [5], it is natural that some will attempt to do the same with fisheye images. In [3], HOG features followed by a support vector machine (SVM) classifier is applied to multiple rotated copies of a fisheye image in 4-degree steps. The process, however, is quite inefficient. Subsequently, [34] uses estimated sizes of people at different locations and foreground segmentation to filter ROIs, speeding up the process to over 10 fps on a regular PC. Multiple SVM classifiers are trained for different radial zones to handle appearance variations. Other than HOG, ACF [8] is another set of features applied to fisheye images [6]. In [4] and [13], the gradient features of HOG and ACF, instead of the image itself, are transformed, and the classifier built for perspective views is used.

CNN based detectors have also been applied to pedestrian images in fisheye images [16,22,31,32,35]. Some of these methods use popular detectors like YOLO without further training, instead choosing to transform the fisheye image to make them suitable for these detectors. For example, [31] creates dozens of perspective views from a single fisheye image and apply YOLO on them individually. Due to the high number of redundant detections from the many overlapping perspective views, the focus of [31] is to compare different approaches of NMS. In [16], where the objective is to do action classification, Mask-RCNN is used to detect people in panoramas transformed from fisheye images.

Some studies follow a different route, choosing to retrain existing detectors so that they can handle the characteristics of fisheye images better. An earlier example is [22] where a simplified version of YOLO-tiny is trained using fisheye images and an extra input channel representing foreground segmentation. The detector works more like a foreground region classifier and fails to detect stationary people. Transfer learning of Mask-RCNN for fisheye images is reported in [35], but the datasets lack the variety usually required for training such a detector. To counter this problem, [32] recently proposes to train the detector using images from the COCO dataset [17] with random rotation. It is also the only existing CNN based detector that outputs rotated bounding boxes and proposes a new NMS scheme called bounding box refinement (BBR).

### III. Methods

*A. Formation of Individual Perspective Views*

There are several geometrical models of image formation for fisheye cameras. The model that is most commonly used in today's commercial fisheye cameras is called the equi-distance model [30], which is explained in the following: A typical fisheye image has a circular shape. Let the center and radius of the circle be $C=(c_x,c_y)$ and $R_0$, respectively. For a point in the fisheye image at distance $R$ ($R<=R_0$) from $C$, its corresponding "line of sight" from the camera center has an angle relative to the camera's optical axis given by

$$\theta = \theta_0 \frac{R}{R_0}. \tag{1}$$

Here $\theta_0$ is part of the camera's spec and is usually close to 90 degrees.

Fig. 2 illustrates the formation of a perspective image from a fisheye image. Here $\boldsymbol{x}_c$, $\boldsymbol{y}_c$, and $\boldsymbol{z}_c$ are the three axes of the camera frame of coordinates. The camera's optical axis is toward $\boldsymbol{z}_c$, and $\boldsymbol{x}_c$ and $\boldsymbol{y}_c$ also represent the axes of the fisheye image frame. And let $\boldsymbol{x}_m$, $\boldsymbol{y}_m$, and $\boldsymbol{z}_m$ be the axes of the perspective image frame, respectively. We specify $\boldsymbol{z}_m$ using two angles, $\varphi_1$ and $\varphi_2$, such that

$$\hat{\boldsymbol{z}}_m = \sin\varphi_1 \cos\varphi_2\, \hat{\boldsymbol{x}}_c + \sin\varphi_1 \sin\varphi_2\, \hat{\boldsymbol{y}}_c + \cos\varphi_1\, \hat{\boldsymbol{z}}_c. \tag{2}$$

The other two axes of the perspective image frame are then given by

$$\hat{\boldsymbol{x}}_m = (\hat{\boldsymbol{z}}_c \times \hat{\boldsymbol{z}}_m)/|\hat{\boldsymbol{z}}_c \times \hat{\boldsymbol{z}}_m| \tag{3}$$

and

$$\hat{\boldsymbol{y}}_m = \hat{\boldsymbol{z}}_m \times \hat{\boldsymbol{x}}_m. \tag{4}$$

Let $\alpha_x$ and $\alpha_y$ be the view angles of the perspective image in horizontal and vertical directions, respectively. A pixel $\boldsymbol{p}=(x_p,y_p)$ in the perspective image corresponds to a ray of direction

$$\boldsymbol{p}_c = \hat{\boldsymbol{z}}_m + x_p \tan(\alpha_x/2)\hat{\boldsymbol{x}}_m + y_p \tan(\alpha_y/2)\hat{\boldsymbol{y}}_m \tag{5}$$

in the camera frame, with both $x_p$ and $y_p$ being relative coordinates in the range of $-1\sim 1$. The corresponding coordinates of $\boldsymbol{p}_c$ in the fisheye image frame are

$$x = (\boldsymbol{p}_c \bullet \hat{\boldsymbol{x}}_c)R + c_x, \quad y = (\boldsymbol{p}_c \bullet \hat{\boldsymbol{y}}_c)R + c_y, \tag{6}$$

where, with $R_0$ and $\theta_0$ specified, we can use (1) to compute $R$ with

$$\theta = \tan^{-1}\left(\frac{\sqrt{(\boldsymbol{p}_c \bullet \hat{\boldsymbol{y}}_c)^2 + (\boldsymbol{p}_c \bullet \hat{\boldsymbol{x}}_c)^2}}{|\boldsymbol{p}_c \bullet \hat{\boldsymbol{z}}_c|}\right). \tag{7}$$

These equations define a geometrical transformation to form a perspective view from the fisheye image. During actual usage, the mappings can be pre-computed and stored in lookup tables to speed up the actual transformation.

*B. Formation of Composite Images*

The size of the composite images can be selected to match the input image sizes suitable for the selected detector. Since we are doing experiments with the YOLO detectors, we set the size of our composite images to 608, the largest commonly used input image size for YOLO. Each composite image is composed of 8 perspective views, also called patches in this paper. These patches are arranged in two rows, and each patch has a size of 152x304. The choice of the number of patches involves a tradeoff: Too few patches and the horizontal view angle of each patch may become too large that pedestrians appear tilted, reducing detection accuracy; too many patches and horizontal view angle of each patch may become too small that most pedestrians are cropped. Overall, we find 8 patches to work well for the datasets.

Without the actual camera specs or calibration information used in the public datasets, we always assume $\theta_0$ is 90 degrees and $C$ is at the image center. $R_0$ is directly estimated from the fisheye images and is kept constant for each dataset. For the individual patches, we set $\alpha_x$ and $\alpha_y$ to 48 and 96 degrees, respectively, with their ratio consistent with the aspect ratio of a patch. The angle $\varphi_1$ is 36 degrees. The 8 perspective views have their $\varphi_2$ in 45-degree steps. These parameters are selected such that the 8 patches together

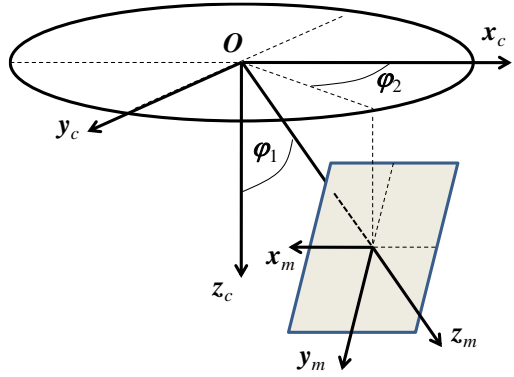

Fig. 2. The geometrical relation between the camera and perspective image frames of coordinates. The camera is at the origin $O$.

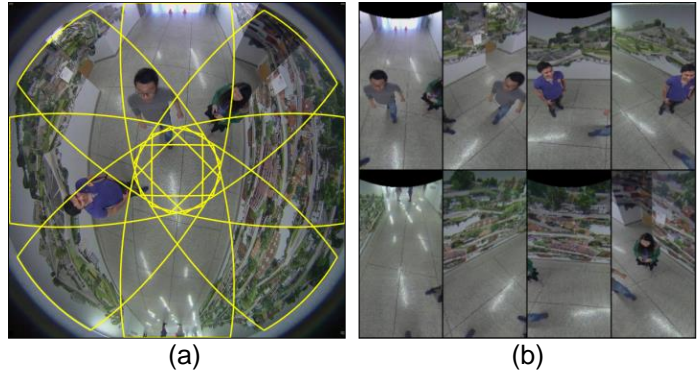

(a)  (b)

Fig. 3. An example of the formation of the composite of perspective views from a fisheye image. (a) The fisheye image overlaid with the regions covered by the perspective views. (b) The composite image consisting of 8 perspective views.

approximately cover the whole fisheye image.

We show in Fig. 3(a) and (b) a fisheye image and a composite image obtained from it, respectively. Regions covered by the 8 perspective views are overlaid on Fig. 3(a). We select the angles such that there's significant overlap around the image center, allowing a person in that region to be completely contained in at least one patch. Some gaps between patch regions are allowed near the outer boundary of the fisheye circle, as persons there are already very small and difficult to detect.

### C. Bounding Box Mapping Exemplars

To get the detections in the fisheye image, bounding boxes generated by the object detector have to be mapped back to the fisheye image frame. We achieve this using a set of mapping exemplars for each patch. Each mapping exemplar consists of one rectangle in the patch (called a *reference box* below) and one rotated rectangle (called a *target box* below) in the fisheye frame. The basic idea is that, for a given detection box in a patch, we identify the mapping exemplars whose reference boxes have high degrees of overlap with the detection box. The target boxes of the selected exemplars are then combined to give a rotated detection box in the fisheye frame. The overall approach can be considered an exemplar-based regression. All the target boxes, as well as the estimated detection boxes in the fisheye frame, are polar-axis-aligned rotated rectangles, meaning that their two axes are along the radial and tangential directions.

To build the set of mapping exemplars, we start with building a set of target boxes in the fisheye frame. Since our focus is on pedestrian detection, we only generate target boxes according to estimated apparent pedestrian sizes in fisheye images. Following the approach in [34], we use cylinders to approximate people in the scene. Given the following parameters: camera height from the ground plane, the height and radius of the cylinder, and the ground-plane location of the cylinder, we can compute the region covered by the person in the fisheye image. This is illustrated in Fig. 4(a) and (b). Each target box is specified by its width, height, and center coordinates, which are determined by the smallest polar-axis-aligned rectangle enclosing the region.

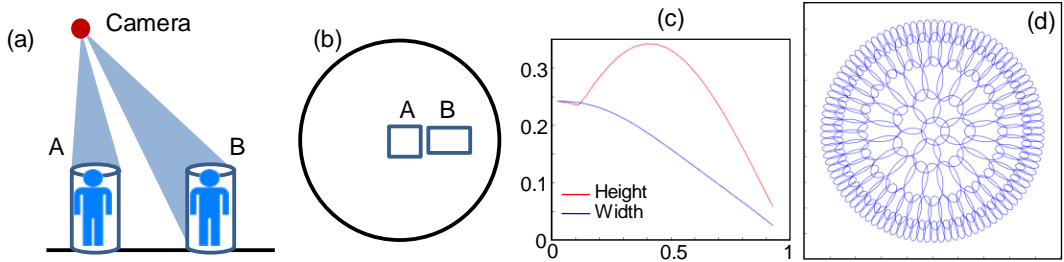

Fig. 4. The process of computing target boxes of the mapping exemplars. (a) The cylindrical model representing typical people. A and B are two people at two different positions. (b) The approximation of A and B in the fisheye image. (c) The height and width of a typical person at different distances from the fisheye image center. Here the scales of both axes are relative to the fisheye image radius. (d) Target boxes plotted in the fisheye frame. For easier viewing, the boxes are plotted as ellipses and a very low overlap ratio (0.2) between adjacent boxes is used.

Due to rotational symmetry, for a given set of parameters, the width and height of a target box are functions of the distance between the center of the box and the image center. In Fig. 4(c) we plot the width and height curves using this set of parameters: camera height: 3 m; person height: 1.7 m; person diameter: 0.5 m. However, since there are a lot of variations of person height and radius resulting from individual differences, poses, and occlusion, we use all combinations of height of 1.3, 1.5, and 1.7 m, and diameter of 0.45 m and 0.6 m. In addition, without the knowledge about the actual camera heights, we use two values, 2.75 m and 3.25 m, to approximate the range typical of indoor environments. These combine to give us 12 sets of parameter. For each set of parameter, target boxes are sampled such that the overlap ratio between adjacent boxes are at least 0.8 in both radial and tangential directions. Fig. 4(d) shows the set of target boxes computed using only one parameter setting (the same as Fig. 4(c)). For clarity of viewing, the boxes are plotted as ellipses and a very low overlap ratio (0.2) between adjacent boxes is used.

Each exemplar $m$ consists of $b_{tgt}(m)$ and $b_{ref}(m)$, its target and reference boxes, respectively. With the target boxes already computed, we need to determine their corresponding reference boxes. This process is done for each patch separately, resulting in a separate set of mapping exemplars, $M(P)$, for patch $P$. As illustrated in Fig. 5, we map all the pixels inside the ellipse enclosed by the target box to the patch frame. The reference box is just the minimal enclosing rectangle of those pixels that are actually inside the patch after mapping. We keep with a mapping exemplar $m$ a "containment ratio" $f_c(m)$, which is the ratio of pixels kept after the mapping here. This gives us a measure of confidence of this mapping exemplar. We discard a mapping exemplar if either of the following two conditions is true: (1) The containment ratio is below 0.1; this means that the mapping is not reliable. (2) The target box is less than 20 pixels in height. For our datasets, the size of $M(P)$ ranges approximately between 8000 and 15000.

### D. Bounding Box Mapping

Typical detections obtained by standard object detectors are rectangular bounding boxes. Since the detection is done on the composite images, it is possible that some generated bounding boxes overlap with multiple patches. We first determine for each bounding box the patch that contains its center, and then crop it to be within that patch.

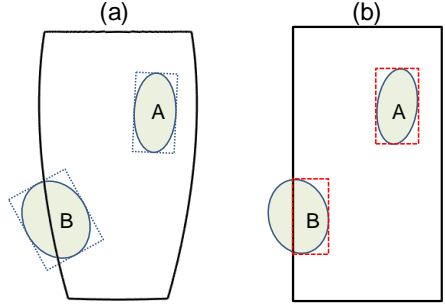

Fig. 5. The illustration of computing reference boxes from target boxes. (a) The region of a patch in the fisheye frame (see Fig. 3). A and B are two examples of target boxes (dashed rectangles) and their enclosed ellipses (shaded). (b) The ellipses of the two boxes in the patch frame (shaded), and the resulting reference boxes (red dashed rectangles).

Let $b^{(p)}$ be a detection bounding box (after cropping) in patch $P$. Our objective here is to compute the corresponding rotated bounding box $b^{(f)}$ in the fisheye frame. A subset of mapping exemplars, denoted $M^*(P,b^{(p)})$, is selected from $M(P)$ to contain only the $k_r$ mapping exemplars with the highest overlap with $b^{(p)}$, defined as

$$f_{ov}(b^{(p)},m) = IOU(b^{(p)}, b_{ref}(m)) \ . \tag{8}$$

Here IOU represents intersection-over-union. We use $k_r$=10 in our experiments.

Given $M^*(P,b^{(p)})$, the detection bounding box in the fisheye frame is computed as the weighted average of the target boxes of the selected mapping exemplars:

$$b^{(f)} = \sum_{m \in M^*(P,b^{(p)})} \lambda(m) b_{tgt}(m) \ . \tag{9}$$

Here $b^{(f)}$ and $b_{tgt}(m)$ are treated as a 4-element vectors in the form of [*center_x center_y width height*]. Since they are in the fisheye frame, their orientations are determined by their center locations relative to $C$. The weighted average is computed separately for the 4 elements. The weights, $\lambda(m)$, are given by

$$\lambda(m) = \frac{f_{ov}(b^{(p)},m)}{\sum_{m' \in M^*(P,b^{(p)})} f_{ov}(b^{(p)},m')} \ . \tag{10}$$

Fig. 6 displays an example of bounding box mapping from the composite image of perspective views (Fig. 6(a)) to the fisheye image (Fig. 6(b)). Of particular interest is the no. 2 box: It is partially cropped in the perspective view, but almost cover the whole person in the fisheye image. This indicates that our mapping method has some degree of ability to recover a partially cropped person.

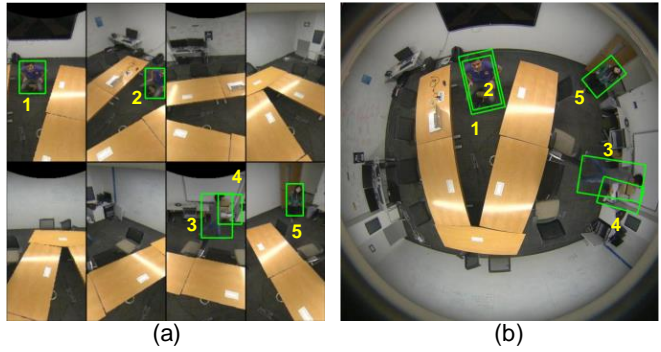

Fig. 6. An example of bounding box mapping. (a) Bounding boxes in the composite image. (b) Bounding boxes in the fisheye image after mapping. The numbers indicate corresponding boxes.

In addition to the bounding box location, we also scale its confidence value based on the "goodness of matching". Specifically, two factors are considered here:

(1) The weighted average of the containment ratios:

$$f_c^*(b^{(p)}) = \sum_{m \in M^*(P, b^{(p)})} \lambda(m) f_c(m). \tag{11}$$

This is a measure of our confidence on the selected target boxes themselves.

(2) The weighted average of the overlap of the detection (in patch frame) with the selected reference boxes:

$$f_{ov}^*(b^{(p)}) = \sum_{m \in M^*(P, b^{(p)})} \lambda(m) f_{ov}(b^{(p)}, m). \tag{12}$$

This is a measure of how well the detection actually represents a person. A low value indicates that the detection might be much smaller or larger than the reference boxes at similar locations, or has an unlikely aspect ratio, and therefore is more likely to be a false detection.

Overall, the confidence value scaling is computed as

$$s(b^{(f)}) = s(b^{(p)}) f_c^*(b^{(p)}) f_{ov}^*(b^{(p)}). \tag{13}$$

Here $s(b^{(p)})$ and $s(b^{(f)})$ are the confidence value before and after scaling, respectively.

*E. Non-maxima Suppression*

NMS (non-maxima suppression) is aimed at consolidating similar detections and reducing false positives. Our implementation of NMS includes two stages. The first stage is done directly on the detection boxes in the patch frame, in the same way NMS is done in common detectors like YOLO. However, we use a much higher IOU threshold (we use 0.8 instead of 0.45 in YOLO). The purpose is only to avoid highly overlapping detections as they basically represent the same object. This step, while optional, can reduce the computational cost in the second-stage NMS. This is because the second stage is done on detection boxes in the fisheye frame, and the computational cost of their IOU is much higher.

The second stage of NMS is done after the detection boxes are mapped to the fisheye frame. We use an exact computation of IOU between rotated rectangles, which is different from the method in [32] where the IOU values are approximated using Monte-Carlo sampling [19]. Three NMS methods are implemented and evaluated in our experiments:

(1) The same NMS as in YOLO with IOU threshold of 0.45.

(2) Gaussian soft NMS proposed in [31]. Instead of directly discarding detections with non-maximal scores, this approach reduces the scores of detections based on the IOU values with neighboring detections that have higher original scores. Starting with an initial set of detections $B_0$, the procedure of adjusting detection scores is given in the following pseudo-code:

$B^* \leftarrow \{\}$ // set of selected detections

While $B_0 \backslash B^* \neq \{\}$

$$b^* = \arg\max_{b \in B_0 \backslash B} s(b)$$

$$B^* \leftarrow B^* \cup \{b^*\}$$

$$s(b') \leftarrow s(b') \exp\left(-IoU(b',b^*)^2 / a_g\right)$$

$$\text{for all } b' \in B_0 \backslash B^*$$

Here $a_g$ is an adjustable parameter, which controls the "strength" of NMS in a way similar to the IOU threshold in regular NMS. This NMS step is applied to axis-aligned bounding boxes in [31], while we process rotated bounding boxes here.

(3) Bounding Box Refinement (BBR) proposed in [32]. Instead of selecting high-score detections, BBR attempts to group nearby detections to form clusters. Mean-shift clustering [2] based on bounding box centers is used with a hyper-spherical binary kernel. Following [32], the radius of the kernel is set to 0.04 of the image size. Once the clustering process converges, the bounding boxes in a cluster are combined into a single one using weighted average, with their scores as the weights. On the other hand, the score of the combined bounding box is the maximum score of the bounding boxes in the cluster.

## IV. EXPERIMENTS

### A. Datasets and Evaluation Protocols

Three public datasets, MW-18Mar [21], PIROPO [23], and Bomni [1,7], are used for evaluating our proposed method. To allow for direct comparison with the results in [32], we use the same subsets of these datasets, including specific video frames and manual annotations, which are provided in [36] by the authors of [32]. We summarize the datasets in Table I.

TABLE I
Overview of Datasets

| Dataset | # Cameras | # Images | # GT Labels |
|---|---|---|---|
| MW-18Mar | 5 | 481 | 1342 |
| PIROPO | 4 | 375 | 803 |
| Bomni | 1 | 331 | 1122 |

TABLE II
Detection Results (AP) with Different Detectors and NMS Methods

| | MW-18Mar | | | | PIROPO | | | | Bomni | | | |
|---|---|---|---|---|---|---|---|---|---|---|---|---|
| | 608-v2 | 416-v2 | 608-v3 | 416-v3 | 608-v2 | 416-v2 | 608-v3 | 416-v3 | 608-v2 | 416-v2 | 608-v3 | 416-v3 |
| YOLO | 0.878 | 0.674 | 0.924 | 0.881 | 0.726 | 0.499 | 0.792 | 0.722 | 0.271 | 0.176 | 0.302 | 0.323 |
| GNMS(0.1) | 0.887 | 0.676 | 0.935 | **0.892** | 0.730 | 0.503 | 0.795 | 0.725 | 0.269 | 0.175 | 0.305 | 0.325 |
| GNMS(0.2) | **0.889** | **0.684** | **0.937** | 0.890 | **0.733** | 0.506 | **0.798** | **0.727** | 0.273 | 0.176 | **0.311** | **0.336** |
| GNMS(0.4) | 0.881 | 0.682 | 0.929 | 0.876 | 0.732 | 0.504 | 0.792 | 0.717 | **0.276** | 0.177 | 0.308 | 0.334 |
| BBR | 0.869 | 0.671 | 0.893 | 0.848 | 0.722 | **0.516** | 0.771 | 0.727 | 0.275 | **0.177** | 0.294 | 0.232 |

Two versions of the well-known YOLO detectors, YOLOv2 [26] and YOLOv3 [27], are used in our experiments. Following [32], we retain all the original detections with scores of at least 0.05 for the *Person* class before the bounding box mapping and NMS steps. The performance is given in terms of average precision (AP) as in PASCAL VOC [10]. When comparing the results with those in [32], we also compute the logarithmic average miss rate (LAMR) proposed in [9]. To compute LAMR, the detection curve is given in miss rate versus false positives per image (FPPI). LAMR is the average of miss rates of 10 FPPI values sampled evenly between 0.01 and one in logarithmic scale.

For all our experiments, two composite images are generated from each fisheye image, with their orientations ($\varphi_2$ in Fig. 2) of respective patches offset by 22.5 degrees. The reported single-image results are always the average results obtained using these two composite images, so as to reduce variations caused by different choices of base orientations.

*B. Pedestrian Detection Results*

The detection results, reported in AP, are listed in Table II. Here we include the results using both YOLOv2 and YOLOv3, as well as two common image sizes, 416 and 608. Each row in Table II corresponds to a different NMS method. Here "YOLO" means that the NMS follows the same criteria as in the original YOLO implementation. "GNMS", meaning Gaussian soft NMS, is the method proposed in [31]. The numbers in the parentheses, 0.1, 0.2 and 0.4, are values of $a_g$ as described in Section III-E. Finally, BBR is the method proposed in [32].

Several observations can be made from the results in Table II. First, YOLOv3 performs significantly better than YOLOv2. This, by itself, is of no surprise. In practice, this means that we can switch the detector when a more suitable one becomes available without retraining, unlike methods in [32,35] that rely on fine-tuning an existing detector for fisheye images. Secondly, larger image size mostly leads to better results. This is different from the results in [32] for calibrated views, but is nonetheless more consistent with what is generally expected. Thirdly, the differences between difference NMS methods are not significant. While GNMS with $a_g$=0.2 seems to perform the best, the results are not too sensitive to the choice of $a_g$. Unless noted otherwise, the subsequent results are obtained using Gaussian NMS with $a_g$=0.2.

Examples of detection results on all 10 camera positions of the three datasets are shown in Fig. 7; the detector used is YOLOv3 with image size of 416. The green, yellow, and red boxes represent true positives, false positives, and missed detections, respectively. The confidence value threshold is 0.2.

Table III is an ablation study of the factors used to scale confidence values (Equation (13)). We can see that, when both factors $f_c^*$ and $f_{ov}^*$ are included, as in (13), the resulting APs are mostly the best, although only slightly. This is probably because only a small portion of detection boxes (mainly those cropped by patch boundaries) are affected by this scaling. For brevity, we only list results using YOLOv2 detectors here. Results from YOLOv3 detectors exhibit similar behaviors.

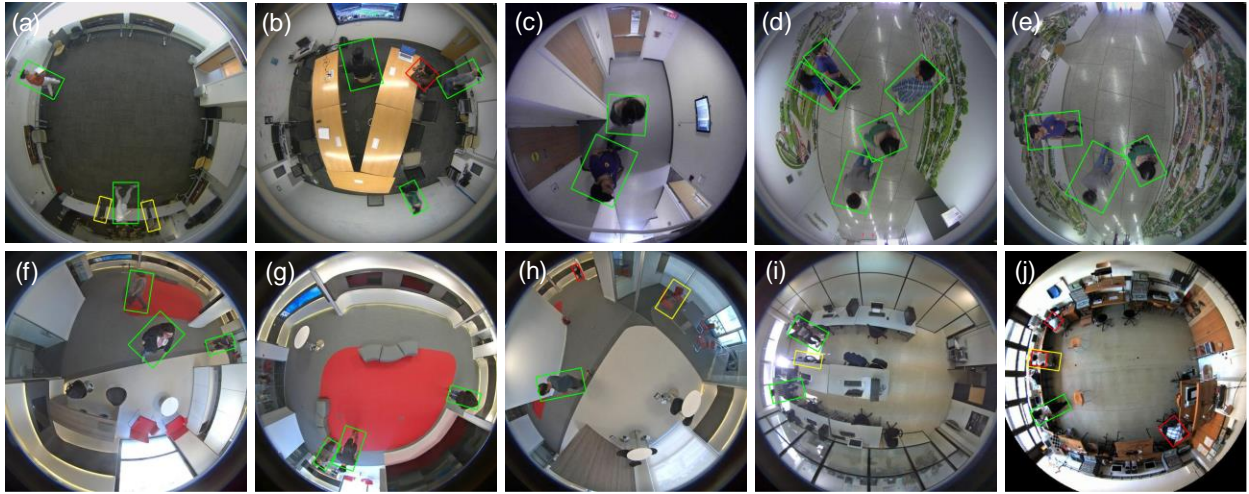

Fig. 7. Example detection results in all the scenes of the datasets ((a)-(e): MW-18Mar; (f)-(i): PIROPO; (j): Bomni). The green, yellow, and red boxes represent true positives, false positives, and missed detections, respectively.

Table IV gives direct comparison of our results with those in [32], so the metric here is LAMR. Our method with Gaussian NMS significantly outperforms [32] in datasets MW-18Mar and Bomni. Even our method with BBR also outperforms [32] in these two datasets. The comparisons are somewhat mixed for PIROPO. Upon further inspection, we find the reason might be related to some "persistent false positives" in the images. This issue will be further discussed in Section V. Even so, our method is still better than [32] in 5 of the 6 settings.

## C. Test-Time Augmentation

It is proposed in [32] that test-time augmentation (integration of detection results from multiple augmented versions of the same source image) can improve detection performance. In [32], 4 copies (combinations of horizontal and vertical flipping) are needed for the results to show improvements. In our experiment here, we just use the two composite images created with an offset in $\varphi_2$ (see Section IV-A), and include all of their detection boxes before NMS. The results listed in Table V show that, using just two composite images (the row labeled as "Ours*2") results in sizable improvements over the original results (the row labeled as "Ours*1"), especially for the PIROPO dataset. For brevity, we only list results using YOLOv2 detectors here. Results from YOLOv3 detectors exhibit similar behaviors.

## D. Timing

The added processing time of transforming images and detections between the fisheye and composite perspective images appears to be of concern for this type of techniques. In our experiments, using a PC with an Intel i7 CPU, we have measured the geometrical transform to take less than 2 ms per image (size of 608x608) using lookup tables. The processing of detections, including bounding box mapping to the fisheye frame and NMS, takes about 3 ms per image on average. The total amount of time above is still much lower than that required of the detector itself (about 30 ms for YOLO on a nVidia 1080Ti GPU). Therefore, we do not consider the added processing time to pose significant problem on performance.

TABLE III
Detection Results (AP) Using Different Confidence Scaling

| Scaling | MW-18Mar | | PIROPO | | Bomni | |
|---|---|---|---|---|---|---|
| | 608-v2 | 416-v2 | 608-v2 | 416-v2 | 608-v2 | 416-v2 |
| None | 0.882 | 0.672 | 0.731 | 0.499 | 0.264 | **0.176** |
| $f_c^*$ | **0.889** | 0.681 | 0.731 | **0.514** | 0.265 | 0.175 |
| $f_{ov}^*$ | 0.876 | 0.675 | 0.727 | 0.489 | 0.269 | **0.176** |
| $f_c^* + f_{ov}^*$ | **0.889** | **0.684** | **0.733** | 0.506 | **0.273** | **0.176** |

TABLE IV
Comparison of Detection Results (LAMR) with Previous Works

| | MW-18Mar | | PIROPO | | Bomni | |
|---|---|---|---|---|---|---|
| | 608-v2 | 416-v2 | 608-v2 | 416-v2 | 608-v2 | 416-v2 |
| Ours+GNMS(0.2) | **0.301** | **0.537** | **0.392** | 0.670 | **0.789** | **0.855** |
| Ours+BBR | 0.325 | 0.550 | 0.407 | 0.659 | 0.790 | 0.861 |
| [32] | 0.361 | 0.558 | 0.407 | **0.639** | 0.887 | 0.893 |

TABLE V
Comparison of Results (AP)
With and Without Test-Time Augmentation

| | MW-18Mar | | PIROPO | | Bomni | |
|---|---|---|---|---|---|---|
| | 608-v2 | 416-v2 | 608-v2 | 416-v2 | 608-v2 | 416-v2 |
| Ours*1 | 0.889 | 0.684 | 0.733 | 0.506 | 0.273 | 0.176 |
| Ours*2 | 0.929 | 0.765 | 0.837 | 0.601 | 0.296 | 0.196 |

## V. Conclusions

In sum, we describe in this paper a new method of detecting pedestrians in top-view fisheye images. While it is not a new idea to first create perspective views from a top-view fisheye image before pedestrian detection, such techniques usually suffer from the higher computational cost due to the large number of perspective views used. Our proposed technique, on the other hand, uses composite images of perspective views to solve this problem. Using just one composite image, we are able to detect pedestrians with good performance with minimal extra computation beyond similar tasks for perspective images. We also present an exemplar based method to map bounding boxes in the perspective views, even ones that are partially cropped, to the fisheye image. For a CNN based detector, it is possible to replace this exemplar based mapping with a branch in the neural network, further improving the speed.

One advantage of our method in practice is that no retraining of the detector is needed. While we can easily switch between YOLOv2 and v3 in our experiments, we can also switch to other detectors, such as the well-known Faster RCNN [28] or SSD [18], with minimal efforts. Furthermore, we can leverage the whole capability of detectors trained on perspective images to detect other object classes, such as in traffic surveillance or room modeling applications. These are definitely directions of future research that we are interested in pursuing.

Since the appearances and sizes of people close to the centers of fisheye images are very different from those of people located in the peripheral regions, the integration of results from multiple detectors specialized for pedestrian detection in different regions is also an approach worth exploring. This relieves a single trained detector from having to detect people of very different appearances and sizes. The integration can be done at the level of detections, such as in [35], or built into the detector network where separate branches are made to specialize on people of different appearances and sizes, more like the approach in [15].

Regarding future studies on pedestrian detection in top-view fisheye images, one of the essential needs is a dataset with sufficient diversity. The three public datasets used in this paper are all designed for tracking tasks, with very few identities of people and even less diversity in terms of the scenes, making them suitable for detector training. A related problem we have observed is that there are some "persistent false positives" (objects that look like people and appear in many frames; see Fig. 7(a), (h) and (i) for examples) and "persistent false negatives" (nearly stationary people that stay in difficult-to-detect poses in many frames; see Fig. 7(b) and (j) for examples) that can significantly skew the evaluated detection performance. As a result, we will also use our experience to build a dataset with much higher diversity than what is currently available. Such a dataset will definitely benefit future research in this area, and performance evaluations will be much more reliable.

## Acknowledgements

This work is supported by the Ministry of Science and Technology of R.O.C. under grant MOST-107-2622-8-009-020.